%% file: main.tex
\documentclass[runningheads]{llncs}

\usepackage{eccv}

\usepackage{eccvabbrv}

\usepackage{graphicx}
\usepackage{booktabs}

\usepackage{bm}
\usepackage{xspace}
\usepackage{multirow}
\usepackage[table]{xcolor}

\definecolor{pinkhl}{HTML}{F7E1E3}
\definecolor{purplehl}{HTML}{E9E3F2}

\usepackage[accsupp]{axessibility}  

\usepackage{hyperref}

\usepackage{orcidlink}

\newcommand{\OurModel}{Allo\{SR\}$^2$\xspace}

\begin{document}

\title{Allo\{SR\}$^2$: Rectifying One-Step Super-Resolution to Stay Real via Allomorphic Generative Flows} 

\titlerunning{\OurModel}

\author{
    Zihan Wang \inst{1}\thanks{Equal contribution. $^{\dagger}$ Corresponding author.} \and 
    Xudong Huang \inst{2}$^{\star}$ \and 
    Junbo Qiao \inst{1} \and \\ 
    Wei Li \inst{2}$^{\dagger}$ \and 
    Jie Hu \inst{2} \and 
    Xinghao Chen \inst{2} \and 
    Shaohui Lin \inst{1, 3}$^{\dagger}$
}

\authorrunning{Z. Wang \etal}

\institute{
    $^1$ East China Normal University \quad
    $^2$ Huawei Foundation Model Dept. \\
    $^3$ Key Laboratory of Advanced Theory and Application in Statistics and Data Science
}

\maketitle

\input{section/00-Abstract}

\input{section/01-Introduction}

\input{section/02-RelatedWork}

\input{section/03-Preliminaries}

\input{section/04-Methodology}

\input{section/05-Experiments}

\input{section/06-Conclusion}


\section*{Acknowledgements}

This work is supported by the National Natural Science Foundation of China (No. 62572193), the Open Research Fund of Key Laboratory of Advanced Theory and Application in Statistics and Data Science, Ministry of Education, and the Fundamental Research Funds for the Central Universities.
We also appreciate Ashin and his brand STAYREAL for inspiring the name of our method.

%
%
\bibliographystyle{splncs04}
\bibliography{section/07-References}
\end{document}

%% file: section/00-Abstract.tex
\begin{abstract}

Real-world image super-resolution (Real-SR) has been revolutionized by leveraging the powerful generative priors from Diffusion Models (DMs) and Flow Matching (FM).
However, existing one-step methods typically replace Gaussian noise with degraded low-resolution (LR) latents at initialization, introducing a substantial distribution shift that further leads to trajectory deviation and prior collapse under extreme acceleration.
To overcome these limitations, we propose \OurModel, a novel FM-based framework that rectifies one-step SR flows via allomorphic generative flows to maintain high-fidelity generative realism.
Specifically, we utilize SNR-Guided Trajectory Initialization to identify a statistically aligned intermediate state along the pre-trained path to integrate LR representations into the generative flow.
To ensure a stable, low-curvature path for one-step inference, we propose Flow-Anchored Trajectory Consistency (FATC), which explicitly regularizes the velocity field of the underlying probability flow.
Furthermore, we develop Allomorphic Trajectory Matching (ATM), a self-adversarial distillation strategy that jointly models the SR flow and the generative flow within a unified velocity field, enabling one-step Real-SR while preserving the generative prior.
Extensive experiments on both synthetic and real-world benchmarks demonstrate that \OurModel achieves state-of-the-art performance in one-step Real-SR, offering a superior balance between fidelity and realism while maintaining extreme efficiency.

\keywords{Super-Resolution \and Generative Prior \and Allomorphic Flow}

\end{abstract}

%% file: section/01-Introduction.tex
\section{Introduction}

Image super-resolution (SR)~\cite{dong2014learning, dong2016image, liang2021swinir, zhang2021designing, wu2024seesr} serves as a fundamental task in low-level vision, aiming to reconstruct high-resolution (HR) images from their low-resolution (LR) observations.
While traditional SR methods~\cite{dong2014learning, dong2016image} achieved remarkable fidelity under predefined synthetic degradations (\eg, bicubic downsampling), they frequently produce over-smoothed results in practical scenarios.
Consequently, the field has shifted toward real-world SR (Real-SR)~\cite{zhang2021designing, wang2021real}, which tackles complex and unknown degradations present in real-world images.

\begin{figure*}[t!]
    \centering
    \includegraphics[width=1.0\linewidth]{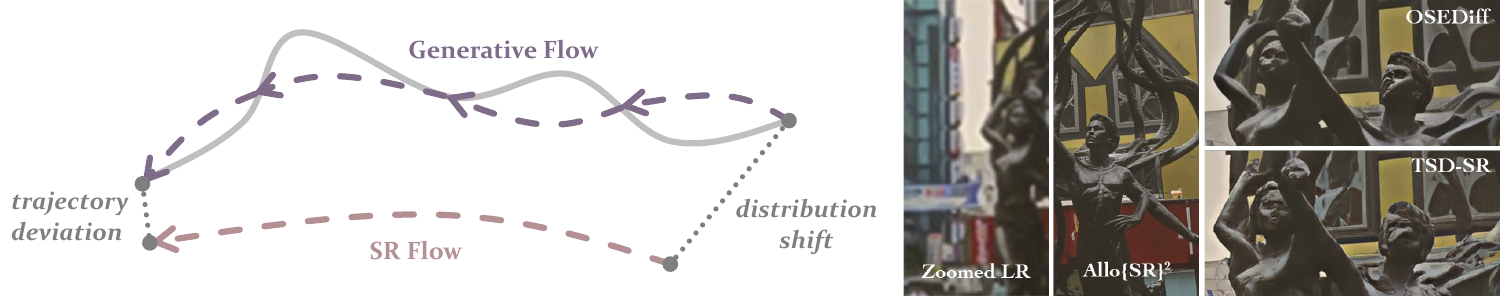}
    \caption{(Left) Existing methods directly replace Gaussian noise with LR latents, leading to a distribution shift and trajectory deviation from the pre-trained generative path. 
    (Right) Compared to other DM- and FM-based one-step methods like OSEDiff~\cite{wu2024one} and TSD-SR~\cite{dong2025tsd}, our approach preserves the generative prior more effectively, recovering realistic textures in severely degraded regions where others suffer from prior collapse.}
    \label{fig:motivation}
\end{figure*}

Early Real-SR methods leveraged Generative Adversarial Networks (GANs)~\cite{goodfellow2014generative, zhu2017unpaired, karras2019style} to recover high-frequency details by pushing the reconstructed outputs toward the natural image manifold.
However, GAN-based methods~\cite{zhang2021designing, wang2021real, liang2022details} are inherently prone to mode collapse and often introduce synthetic artifacts.
Recently, Diffusion Models (DMs)~\cite{song2019generative, ho2020denoising, song2021score, peebles2023scalable, nichol2021improved} and Flow Matching (FM) paradigms~\cite{lipman2023flow, albergo2025stochastic, albergo2023building, liu2022flow, liu2024instaflow} have revolutionized image generation, demonstrating unprecedented capabilities in synthesizing realistic visuals by progressively transforming Gaussian noise into high-quality images. 
Driven by this success, an emerging trend in Real-SR adapts these pre-trained DM- and FM-based models as powerful generative priors~\cite{wang2024exploiting, lin2024diffbir, wu2024seesr, yang2024pixel}, leveraging their rich, encapsulated knowledge of natural image statistics to reconstruct highly realistic HR details.
Nevertheless, their inherent iterative sampling process typically requires tens to hundreds of function evaluations to synthesize a single image, resulting in prohibitive computational latency for practical deployment.

To break this efficiency bottleneck, numerous attempts~\cite{yue2023resshift, zhang2025uncertainty, wang2024sinsr, you2025consistency, xu2025fast, wu2024one, dong2025tsd, sun2025pixel} have been made to compress these continuous-time trajectories into a few steps or even one step.
For instance, ResShift~\cite{yue2023resshift} and its extension~\cite{zhang2025uncertainty} propose replacing Gaussian noise with noised LR images as the initial state of the diffusion process, thereby significantly shortening the required reverse path.
SinSR~\cite{wang2024sinsr} further reformulates this framework as an ordinary differential equation (ODE) and directly distills it into one step.
Some other works~\cite{you2025consistency, xu2025fast} extend Consistency Models to Real-SR by enforcing consistency along the generative trajectory to enable one-step reconstruction.
Beyond these approaches, another active line of research explores Score Distillation~\cite{wu2024one, dong2025tsd, sun2025pixel, you2025consistency}, which aligns the reconstructed image distribution with the pre-trained generative prior by minimizing the Kullback–Leibler (KL) divergence.

While these few-step and one-step approaches significantly improve efficiency, seamlessly transferring pre-trained generative prior to one-step Real-SR under such extreme acceleration remains a critical challenge.
As illustrated in Fig.~\ref{fig:motivation} (left), existing methods typically adapt generative models by directly replacing the initial Gaussian noise with degraded LR latent representations.
This naive substitution induces a substantial \emph{distribution shift}, as the latent distribution of LR images is fundamentally misaligned with the Gaussian prior expected by the pre-trained network.
Under one-step inference, this misalignment, compounded by the absence of explicit velocity field constraints, leads to a severe \emph{trajectory deviation}.
Consequently, the model suffers from \emph{prior collapse}, where it fails to fully exploit its generative prior, resulting in either over-smoothed outputs or unnatural artifacts, as shown in Fig.~\ref{fig:motivation} (right).
Although recent efforts~\cite{yue2025arbitrary, wu2025omgsr} attempt to mitigate this distribution shift via learnable noise injection or VAE encoder fine-tuning to align LR latents with intermediate states of the generative flow, they still treat the underlying generative model merely as a static backbone.
We argue that rather than treating the pre-trained network as an isolated backbone, the SR task should integrate it as a dynamic anchor within a unified flow framework.
Our key insight is that image generation and restoration share an \emph{allomorphic} nature: they represent structurally analogous trajectories converging toward the same target manifold of realistic HR images.
While the generative flow provides an optimal, distortion-free trajectory, the SR flow serves as its localized, condition-constrained counterpart.

Inspired by this observation, we propose a novel FM-based Real-SR method, called \textbf{\OurModel}, which rectifies one-step \textbf{S}uper-\textbf{R}esolution to \textbf{S}tay \textbf{R}eal via \textbf{Allo}morphic generative flows.
First, we rethink the injection point of the LR representations from a mathematical standpoint.
Through Signal-to-Noise Ratio (SNR) analysis, we demonstrate that the LR latent features exhibit the highest statistical alignment with an intermediate state along the pre-trained generative trajectory.
Consequently, this state naturally serves as the optimal initialization for the local SR flow.
Instead of standard conditioned fine-tuning, we construct an asymmetric dual-path framework that concurrently models a global generative flow (from noise to HR) and a local SR flow (from LR to HR).
By sharing weights and enforcing a cross-trajectory score matching constraint between corresponding timesteps, the pre-trained generative flow serves as a dynamic anchor.
It continuously transfers its distortion-free prior to rectify the SR flow, effectively preventing prior collapse and explicitly matching the distribution of SR outputs to the natural image manifold.
Remarkably, this structural rectification mitigates over-smoothed reconstructions while minimizing trajectory curvature, flattening the integration path to enable high-fidelity, photorealistic Real-SR in a single function evaluation.

Our main contributions are summarized as follows: 
\begin{itemize}
    \item We propose \OurModel, a novel FM-based framework that rectifies one-step SR flow through the allomorphic nature between generation and restoration.
    \item We utilize SNR-Guided Trajectory Initialization paired with Flow-Anchored Trajectory Consistency (FATC) to mitigate the distribution shift and trajectory deviation, ensuring a stable, low-curvature SR flow.
    \item We develop Allomorphic Trajectory Matching (ATM), a self-adversarial distillation strategy that effectively transfers generative prior to the SR flow, preventing prior collapse while preserving perceptual realism.
\end{itemize}

%% file: section/02-RelatedWork.tex
\section{Related Work}

\subsection{Real-World Image Super-Resolution}

Image super-resolution (SR)~\cite{dong2014learning, dong2016image, liang2021swinir, zhang2021designing, wu2024seesr} aims to recover high-resolution (HR) details from degraded low-resolution (LR) observations.
Early SR methods~\cite{dong2014learning, dong2016image} primarily utilize simple synthetic degradations to generate training pairs, which often fail to generalize to complex and stochastic degradations in real world.

To bridge this gap, Generative Adversarial Networks (GANs)~\cite{goodfellow2014generative, zhu2017unpaired, karras2019style} were introduced to perform real-world SR (Real-SR) with more complex degradations.
Pioneering work~\cite{ledig2017photo} demonstrates that combining adversarial objectives with perceptual losses yields superior visual quality compared to traditional pixel-wise constraints.
Building upon this, subsequent research further improves the realism and diversity of synthesized training data through explicit degradation modeling.
Specifically, BSRGAN~\cite{zhang2021designing} employs randomly shuffled degradation operators to better approximate real-world conditions,
while Real-ESRGAN~\cite{wang2021real} further extends this paradigm via a high-order degradation modeling process.
Despite these advancements, GAN-based Real-SR methods still suffer from the instability of adversarial training and the risk of mode collapse, which are highly prone to generating unnatural visual artifacts. 
Although several refined approaches~\cite{liang2022details, zhang2022perception, wang2019deep} partially mitigate these issues, the inherent limitations of GANs continue to constrain the synthesis of highly natural and photorealistic details, ultimately bounding their generative capacity.

With the success of Diffusion Models (DMs)~\cite{ho2020denoising, song2021score}, recent studies have extensively harnessed generative priors to  address the Real-SR task.
To leverage the pre-trained Stable Diffusion model~\cite{rombach2022high}, StableSR~\cite{wang2024exploiting} and DiffBIR~\cite{lin2024diffbir} inject LR representations as conditional guidance via specialized feature modulation, successfully recovering high-frequency details.
Beyond low-level conditioning, semantic information has also been incorporated to enhance restoration quality.
For example, SeeSR~\cite{wu2024seesr} and PASD~\cite{yang2024pixel} introduce semantic guidance and pixel-aware cross-attention, respectively, enabling more effective modeling of fine-grained structures and degradation-insensitive features.
Meanwhile, ResShift~\cite{yue2023resshift} improves sampling efficiency by reformulating the diffusion process in the residual between HR and LR images.
Nevertheless, these DM-based methods are fundamentally constrained by the iterative reverse denoising process, resulting in prohibitive inference latency compared to GAN-based methods.

\subsection{Acceleration of Diffusion and Flow Matching}

Diffusion Models (DMs)~\cite{song2019generative, ho2020denoising, song2021score, peebles2023scalable, nichol2021improved} have achieved remarkable success across various generative tasks by utilizing a parameterized Markov chain to bridge the natural image manifold with the standard Gaussian prior.
Building upon this continuous-time formulation, the Flow Matching (FM) paradigm~\cite{lipman2023flow, albergo2025stochastic, albergo2023building} learns a time-dependent velocity field to establish deterministic probability flows between distributions.
Recent variants, such as Rectified Flow~\cite{liu2022flow, liu2024instaflow}, further straighten the interpolation trajectories to improve sampling efficiency.

Despite their strong generative performance, the iterative sampling process necessitates tens to hundreds of function evaluations during inference, resulting in substantial computational overhead.
This has motivated extensive research on accelerating the sampling process.
Early efforts, such as Progressive Distillation~\cite{meng2023distillation, salimans2022progressive}, iteratively halve the inference steps of student models through a multi-stage distillation.
However, the compounding errors across these sequential distillation stages may lead to degraded generation quality in the final student model. 
Meanwhile, Consistency Models~\cite{luo2023latent, song2023consistency, kim2024consistency, yang2024consistency} enable one-step or few-step generation by enforcing self-consistency along the sampling trajectory, ensuring that intermediate states map to a shared origin. 
More recently, Score Distillation approaches~\cite{wang2023prolificdreamer, yin2024one, yin2024improved} leverage pre-trained diffusion models as implicit score functions to supervise a one-step generator by minimizing the Kullback–Leibler (KL) divergence between the generated distribution and the target distribution. 
Expanding this paradigm, a recent advancement~\cite{cheng2026twinflow} introduces a dual-flow matching framework that removes the need for auxiliary networks, thereby reducing memory overhead and enabling scalable training on large models.

Driven by these advancements, an emerging line of research seeks to adapt these highly efficient frameworks to the Real-SR task.
SinSR~\cite{wang2024sinsr} reformulates the diffusion trajectory into a deterministic sampling process, distilling the inference process into a single step.
Alternatively, OSEDiff~\cite{wu2024one} leverages Variational Score Distillation (VSD)~\cite{wang2023prolificdreamer, yin2024one} to inherit the generative capacity of pre-trained multi-step prior within a one-step generator.
To mitigate the artifacts and instability caused by VSD during early training stages, TSD-SR~\cite{dong2025tsd} introduces Target Score Distillation (TSD), which improves optimization stability and leads to more robust convergence.
Beyond Score Distillation, CTMSR~\cite{you2025consistency} formulates a deterministic PF-ODE mapping from noisy LR to HR images and employs consistency training combined with a distribution trajectory matching loss to achieve high-fidelity one-step generation. 

%% file: section/03-Preliminaries.tex
\section{Preliminaries}

\subsection{Flow Matching}

Flow Matching provides an intuitive paradigm for generative modeling, aiming to learn a time-dependent velocity field that defines a probability path between a complex target distribution and a simple prior (\eg, a standard Gaussian).
Formally, given a real data sample $\bm{x} \sim p_{\mathrm{data}}(\bm{x})$ and a noise sample $\bm{\epsilon} \sim \mathcal{N}(\bm{0}, \bm{I})$, the probability path is typically parameterized as:
\begin{equation}
    \bm{x}_t = a_t \bm{x} + b_t \bm{\epsilon}, \quad t \in [0,1]
\end{equation}
where $a_t$ and $b_t$ are predefined scalar schedules satisfying boundary conditions $a_0 = 1, b_0 = 0$ and $a_1 = 0, b_1 = 1$.
The corresponding velocity field $\bm{v}_t$ is defined as the time derivative of the path:
\begin{equation}
    \bm{v}_t = \frac{\mathrm{d} \bm{x}_t}{\mathrm{d} t} = a'_t \bm{x} + b'_t \bm{\epsilon}.
\end{equation}
This represents the conditional velocity, denoted as $\bm{v}_t(\bm{x}_t | \bm{x})$. 
In particular, adopting the linear schedule $a_t = 1 - t$ and $b_t = t$ yields a straight-line interpolation between $\bm{x}$ and $\bm{\epsilon}$.
Under this parameterization, the conditional velocity reduces to the time-independent form $\bm{v}_t = \bm{\epsilon} - \bm{x}$.

Since an intermediate state $\bm{x}_t$ does not uniquely identify its originating data sample $\bm{x}$, Flow Matching essentially models the expectation over all possibilities, known as the marginal velocity $\bm{v}(\bm{x}_t, t) = \mathbb{E}[\bm{v}_t(\bm{x}_t|\bm{x})|\bm{x}_t]$.
Directly optimizing this marginal velocity is intractable due to the unknown data density $p_{\mathrm{data}}(\bm{x})$.
Conditional Flow Matching (CFM)~\cite{lipman2023flow} avoids this difficulty by minimizing a tractable surrogate loss between a neural network $\bm{v}_\theta$ and the conditional velocity:
\begin{equation}
    \mathcal{L}_{\mathrm{CFM}} = \mathbb{E} \left[ \left\| \bm{v}_\theta(\bm{x}_t, t) - \bm{v}_t(\bm{x}_t | \bm{x}) \right\|^2 \right].
\end{equation}
Crucially, minimizing $\mathcal{L}_{\mathrm{CFM}}$ is theoretically equivalent to matching the intractable marginal velocity field $\bm{v}(\bm{x}_t, t)$, thereby enabling $\bm{v}_\theta$ to accurately govern the distribution transformation during inference.

\subsection{Score Distillation}

To alleviate the costly iterative sampling of DM- and FM-based models, Score Distillation has emerged as a widely adopted paradigm for one-step or few-step acceleration.
The core principle is to distill the rich distributional knowledge from a pre-trained multi-step teacher model into a one-step generator $G_\theta$.

Mathematically, let $p_\theta(\bm{x})$ denote the distribution generated by the one-step generator, and $p_{\mathrm{data}}(\bm{x})$ denote the target data distribution captured by the multi-step teacher model. 
Recent Score Distillation methods, such as Distribution Matching Distillation (DMD)~\cite{yin2024one, yin2024improved}, seek to align these two distributions by minimizing the Kullback-Leibler (KL) divergence:
\begin{equation}
    \min_\theta \mathcal{D}_{\mathrm{KL}} \left( p_\theta(\bm{x}) \| p_{\mathrm{data}}(\bm{x}) \right).
\end{equation}
Directly evaluating this objective requires access to the underlying probability densities, which are generally unavailable in high-dimensional spaces. Fortunately, optimizing the generator only requires the gradient of the divergence with respect to the generator parameters $\theta$:
\begin{equation}
    \nabla_\theta \mathcal{D}_{\mathrm{KL}} = \mathbb{E} \left[ \Big( \nabla_{\bm{x}} \log p_\theta(\bm{x}) - \nabla_{\bm{x}} \log p_{\mathrm{data}}(\bm{x}) \Big) \frac{\partial G_\theta}{\partial \theta} \right].
\end{equation}
Here, $\nabla_{\bm{x}} \log p_{\mathrm{data}}(\bm{x})$ is the score of the target distribution, which is typically approximated using the teacher model pre-trained on real data.
Meanwhile, $\nabla_{\bm{x}} \log p_\theta(\bm{x})$ represents the score of the generated distribution and is estimated by an auxiliary network trained on fake samples produced by the generator $G_\theta$.
By minimizing the discrepancy between these two score functions, Score Distillation aligns the generated distribution with the target distribution directly.
Consequently, the student generator can inherit the teacher's generative capability without explicitly reproducing the multi-step sampling trajectory, enabling high-quality generation in a single forward pass.

%% file: section/04-Methodology.tex
\section{Methodology}

\begin{figure*}[t!]
    \centering
    \includegraphics[width=\linewidth]{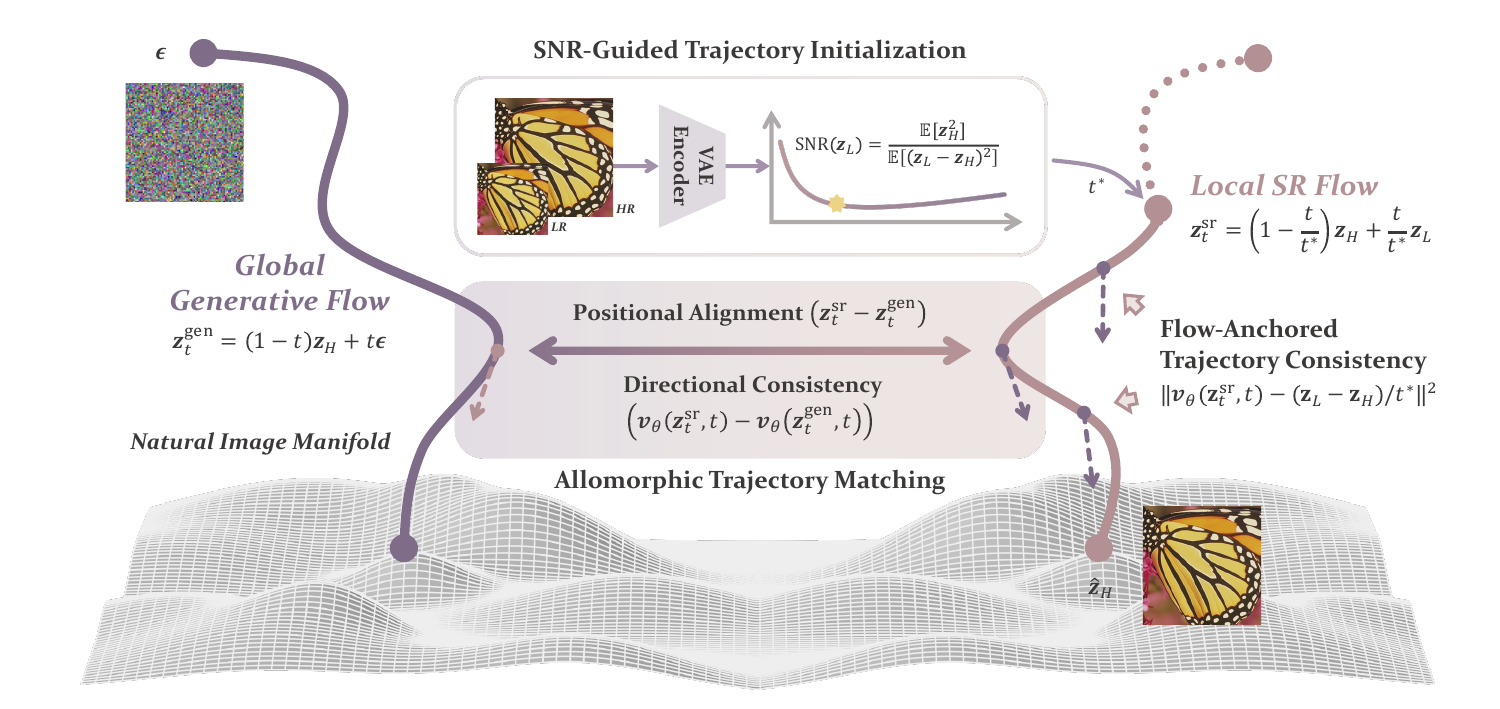}
    \caption{\textbf{Overview of \OurModel.} 
    (a) SNR-Guided Trajectory Initialization identifies the optimal initializing timestep to mitigate the distribution shift between LR latents and the Gaussian prior.
    (b) Flow-Anchored Trajectory Consistency (FATC) enforces a low-curvature SR flow via supervision on the instantaneous velocity. 
    (c) Allomorphic Trajectory Matching (ATM) matches the scores from the local SR flow and the global generative flow, ensuring the SR outputs stay within the natural image manifold.}
    \label{fig:framework}
\end{figure*}

In this section, we present \OurModel, as illustrated in Fig.~\ref{fig:framework}.
Our framework is motivated by the allomorphic nature between generation and restoration, where the SR flow can be viewed as a localized, condition-constrained counterpart of the generative flow.
We first identify the optimal embedding timestep via SNR analysis to integrate LR representations into the pre-trained generative flow (Sec.~\ref{sec:snr}).
To mitigate the trajectory deviation, we introduce a Flow-Anchored Trajectory Consistency (FATC) to regularize the velocity field of the underlying probability flow (Sec.~\ref{sec:fatc}).
Furthermore, we propose Allomorphic Trajectory Matching (ATM) to align the local SR flow with the global generative flow, thereby enabling efficient one-step SR while preserving the generative prior (Sec.~\ref{sec:atm}).

\subsection{SNR-Guided Trajectory Initialization}
\label{sec:snr}

In standard image generation, a continuous-time trajectory transports a noise sample $\bm{\epsilon}$ to the HR latent $\bm{z}_H$.
In contrast, the LR latent $\bm{z}_L$, as a degraded state of $\bm{z}_H$, exhibits a substantial distribution shift from the Gaussian prior.
Directly replacing the initial noise with $\bm{z}_L$ may disrupt the pre-trained generative prior. 
To reconcile this, we interpret $\bm{z}_L$ as an intermediate state composed of the HR latent $\bm{z}_H$ and the degradation residual $\Delta \bm{z} = \bm{z}_L - \bm{z}_H$. 
This motivates identifying an optimal embedding timestep where the theoretical noise level of the pre-trained trajectory best matches the degradation severity of $\bm{z}_L$.

Following recent advances~\cite{wu2025omgsr}, we adopt the Signal-to-Noise Ratio (SNR) as a metric to evaluate latent similarity, as it naturally connects the physical degradation of the image to the decreasing noise schedule of the generative trajectory. 
Given a dataset of $N$ paired LR-HR latents $\{(\bm{z}_L^{(i)}, \bm{z}_H^{(i)})\}_{i=1}^N$ encoded by the fixed pre-trained VAE encoder, we formulate the degraded SNR as the ratio of the signal variance to the residual variance. 
The optimal embedding timestep is then determined by searching over the schedule $t \in \{1, \dots, T\}$ to minimize the discrepancy between the theoretical trajectory SNR and the degraded SNR:
\begin{equation}
    t^* = \arg\min_{t} \frac{1}{N} \sum_{i=1}^N \left| \frac{(1-t)^2 \cdot \mathbb{E}[\|\bm{z}_H^{(i)}\|^2]}{t^2} - \frac{\mathbb{E}[\|\bm{z}_H^{(i)}\|^2]}{\mathbb{E}[\|\bm{z}_L^{(i)} - \bm{z}_H^{(i)}\|^2]} \right|.
\end{equation}

By explicitly initializing the SR trajectory at $t^*$, we establish a physically grounded starting state that directly bypasses the highly noisy integration region $t \in (t^*, 1]$. 
Crucially, this initialization forces the subsequent SR flow to closely adhere to the original generative velocity field, thereby preventing catastrophic forgetting of the pre-trained prior during fine-tuning.

\subsection{Flow-Anchored Trajectory Consistency}
\label{sec:fatc}

Once the optimal initial timestep $t^*$ is determined, our framework can reconstruct HR latent in a single functional evaluation.
Specifically, we formulate the one-step reverse integration as a first-order Euler approximation:
\begin{equation}
\bm{\hat{z}}_H = \bm{z}_L - t^* \bm{v}_\theta(\bm{z}_L, t^*),
\label{eq:reconstruct}
\end{equation}
where $\bm{v}_\theta(\bm{z}_L, t^*)$ denotes the predicted velocity field at the starting state.
To supervise the training process, we employ a reconstruction loss that integrates a latent-wise MSE loss and a pixel-wise perceptual loss based on LPIPS:
\begin{equation}
    \mathcal{L}_{\mathrm{Rec}} = \mathcal{L}_{\mathrm{MSE}}(\bm{\hat{z}}_H, \bm{z}_H) + \lambda \mathcal{L}_{\mathrm{LPIPS}}(D(\bm{\hat{z}}_H), D(\bm{z}_H)).
\end{equation}

Although Eq.~\ref{eq:reconstruct} facilitates a direct mapping, training solely on endpoint objectives often forces the network to fit a highly complex, non-linear function via a shortcut.
In the absence of explicit supervision on the instantaneous velocity field of the underlying probability flow, the learned trajectory is prone to significant deviation from the optimal generative path.
To mitigate this, we introduce Flow-Anchored Trajectory Consistency (FATC) by interpolating between $\bm{z}_H$ and $\bm{z}_L$.
We define the rescaled intermediate state $\bm{z}_t^{\mathrm{sr}}$ as:
\begin{equation}
    \bm{z}_t^{\mathrm{sr}} = \left( 1 - \frac{t}{t^*} \right) \bm{z}_H + \frac{t}{t^*} \bm{z}_L, \quad t \in [0, t^*].
\end{equation}
Since $\bm{z}_L$ represents the state specifically at $t^*$, the interpolation schedule is normalized by $t^*$ to ensure temporal alignment with the pre-trained flow.
During training, we anchor the network to the underlying flow by minimizing the velocity discrepancy across these intermediate states:
\begin{equation}
    \mathcal{L}_{\mathrm{FATC}} = \mathbb{E} \left\| \bm{v}_\theta(\bm{z}_t^{\mathrm{sr}}, t) - \frac{\bm{z}_L - \bm{z}_H}{t^*} \right\|^2.
    \label{eq:fatc}
\end{equation}

By enforcing this velocity-level supervision, the SR model adheres to a low-curvature, linear trajectory. This consistency maximizes the fidelity of the one-step approximation and ensures the SR flow remains within the support of the pre-trained probability density.

\subsection{Allomorphic Trajectory Matching}
\label{sec:atm}

While the FATC in Eq.~\ref{eq:fatc} regularizes the velocity field, fine-tuning pre-trained generative model on limited LR-HR data pairs often precipitates prior collapse.
In this regime, the model sacrifices its generative richness to overfit the specific degradations of the training pairs.
To circumvent this, we propose Allomorphic Trajectory Matching (ATM).
Rather than adapting the generative prior to the SR task, ATM aligns the SR trajectory with the intrinsic generative flow.

Inspired by the self-adversarial framework of TwinFlow~\cite{cheng2026twinflow}, we construct two allomorphic, parameter-shared trajectories within a unified velocity field: a generative flow and an SR flow. 
By sharing weights across both tasks, the model is forced to learn a unified latent representation where SR transitions are perceived as a subset of the broader generative probability flow.
To ensure the SR results remain within the support of the natural image manifold, we treat the generative flow as a dynamic positive guidance.
We minimize the Kullback-Leibler (KL) divergence between the distribution of reconstructed SR latents $p_{\mathrm{sr}}$ and the prior distribution $p_{\mathrm{gen}}$ maintained by the generative flow:
\begin{equation}
    \mathcal{L}_{\mathrm{ATM}} = \mathcal{D}_{\mathrm{KL}} \left( p_{\mathrm{sr}} \parallel p_{\mathrm{gen}} \right) = \mathbb{E} \left[ \log p_{\mathrm{sr}}(\bm{\hat{z}}_H) - \log p_{\mathrm{gen}}(\bm{\hat{z}}_H) \right].
\end{equation}
Following the Score Distillation paradigm, the gradient of $\mathcal{L}_{\mathrm{ATM}}$ is approximated via the score discrepancy between the allomorphic paths:
\begin{equation}
\label{eq:atm}
\begin{aligned}
    \nabla_\theta \mathcal{L}_{\mathrm{ATM}} 
    &= \mathbb{E} \left[ \left( \nabla_{\bm{\hat{z}}_H} \log p_{\mathrm{sr}}(\bm{\hat{z}}_H) - \nabla_{\bm{\hat{z}}_H} \log p_{\mathrm{gen}}(\bm{\hat{z}}_H) \right) \frac{\partial \bm{\hat{z}}_H}{\partial \theta} \right] \\
    &= \mathbb{E} \left[ w(t) \left[ \underbrace{\left(\bm{z}_t^{\mathrm{sr}} - \bm{z}_t^{\mathrm{gen}}\right)}_{\text{Position}} - \underbrace{t \left( \bm{v}_\theta(\bm{z}_t^{\mathrm{sr}}, t) - \bm{v}_\theta(\bm{z}_t^{\mathrm{gen}}, t) \right)}_{\text{Direction}} \right] \frac{\partial \bm{\hat{z}}_H}{\partial \theta} \right],
\end{aligned}
\end{equation}
where $w(t)$ is a time-dependent weighting function. 
since the SR schedule is rescaled by $t^*$ in Eq.~\ref{eq:fatc}, the intermediate states $\bm{z}_t^{\mathrm{sr}}$ and $\bm{z}_t^{\mathrm{gen}}$ exhibit theoretical proximity.
As derived in Eq.~\ref{eq:atm}, the first term performs positional alignment by minimizing the distance between allomorphic states, while the second term ensures directional consistency.
By minimizing $\mathcal{L}_{\mathrm{ATM}}$, we achieve a better alignment between SR outputs and the natural image distribution.

This derivation demonstrates that by matching the scores of the SR flow to the generative flow, we effectively perform self-adversarial alignment.
The SR task is no longer a simple regression; it becomes a specialized navigation within the generative manifold. 
Consequently, the shared weights learn to bridge the degradation gap while preserving global generative capabilities, achieving a superior balance between fidelity and realism.

%% file: section/05-Experiments.tex
\section{Experiments}

\subsection{Experimental Settings}

\subsubsection{Training Datasets}
Following \cite{wu2024seesr, wu2024one}, we train our model on LSDIR~\cite{li2023lsdir} and the first 10K face images from FFHQ~\cite{karras2019style}, totaling 95K images.
The degradation pipeline of Real-ESRGAN~\cite{wang2021real} is used to synthesize LR-HR training pairs.

\subsubsection{Testing Datasets}
We evaluate our method on a synthetic dataset DIV2K-Val~\cite{agustsson2017ntire} and three real-world datasets, including RealSR~\cite{cai2019toward}, DRealSR~\cite{wei2020component} and RealLQ250~\cite{ai2024dreamclear}.
We crop $512 \times 512$ patches from DIV2K-Val and apply the Real-ESRGAN degradation pipeline to downsample them to $128 \times 128$.
For RealSR and DRealSR, which provide paired GT images, we apply center-cropping to the LR images to a resolution of $128 \times 128$.
Since RealLQ250 lacks corresponding GT images, we perform no cropping and maintain the original $256 \times 256$ resolution.
All experiments are conducted with the scaling factor of $\times 4$. 

\subsubsection{Compared Methods}
We compare \OurModel with other DM- and FM-based models, including multi-step models: StableSR~\cite{wang2024exploiting}, SeeSR~\cite{wu2024seesr}, ResShift~\cite{yue2023resshift}; and one-step models: SinSR~\cite{wang2024sinsr}, OSEDiff~\cite{wu2024one}, TSD-SR~\cite{dong2025tsd}, and CTMSR~\cite{you2025consistency}.

\subsubsection{Evaluation Metrics}
To comprehensively evaluate the performance of all methods, we employ a series of full-reference (FR) and no-reference (NR) metrics.
For FR evaluation, PSNR and SSIM~\cite{wang2004image} reflect fidelity, while LPIPS~\cite{zhang2018unreasonable} and DISTS~\cite{ding2020image} measure perceptual quality.
FID~\cite{heusel2017gans} further evaluates the distance of distributions between GT and reconstructed images.
For NR evaluation, we adopt four blind image quality assessment metrics, namely NIQE~\cite{zhang2015feature}, MUSIQ~\cite{ke2021musiq}, MANIQA~\cite{yang2022maniqa}, and CLIPIQA~\cite{wang2023exploring}, to evaluate visual naturalness and structural integrity of the images.

\subsubsection{Implementation Details}
We utilize the pre-trained weights of FLUX.1-dev\footnote{\url{https://github.com/black-forest-labs/flux}} as our foundation generative model.
To facilitate efficient parameter adaptation, LoRA~\cite{hu2022lora} is integrated into both the VAE encoder and the Diffusion Transformer (DiT) components with a rank of $r=64$. 
The framework is optimized using the AdamW optimizer~\cite{kingma2014adam} with a learning rate of $5 \times 10^{-5}$ and a global batch size of 16, for a total of 10K iterations.

\subsection{Comparison Results}

\input{table/Comparison}

\begin{figure*}[t!]
    \centering
    \includegraphics[width=\linewidth]{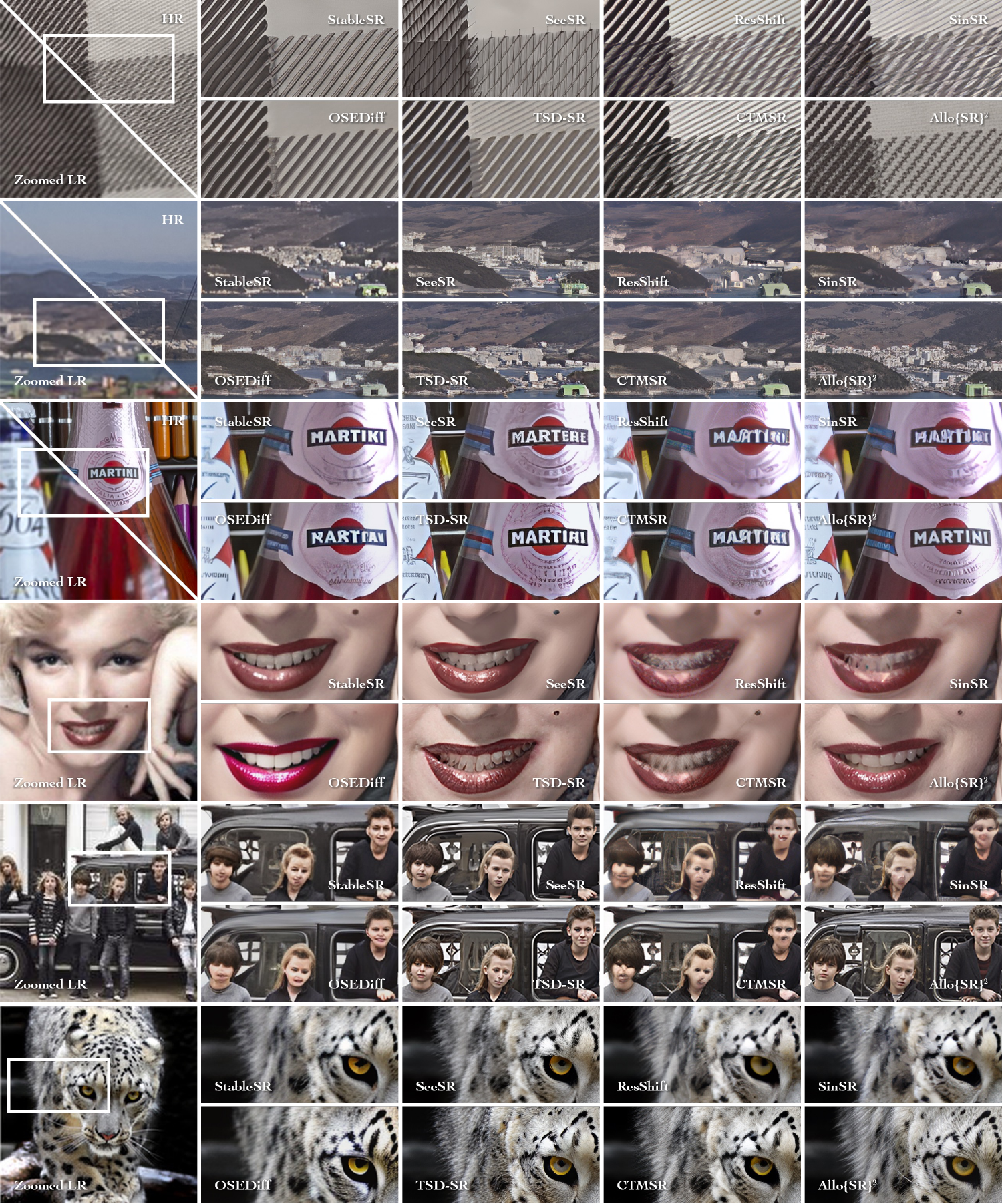}
    \caption{Qualitative comparisons of different DM- and FM-based Real-SR methods.}
    \label{fig:visualization}
\end{figure*}

\subsubsection{Quantitative Comparisons.}

As shown in Tab.~\ref{tab:comparison}, \OurModel achieves an excellent balance between fidelity and realism across both multi-step and one-step methods.
It achieves the best performance on most NR metrics across all four benchmarks.
In particular, it significantly outperforms other one-step methods on MANIQA, which highly correlates with human visual preference. 
Moreover, the superior FID demonstrates that the proposed ATM effectively aligns the restored distribution with the natural image manifold.
Although \OurModel does not achieve the highest scores on PSNR and SSIM, this is consistent with the well-known perception-distortion trade-off in Real-SR~\cite{blau2018perception}.
Since these two metrics emphasize pixel-wise correspondence, they tend to penalize realistic textures while favoring over-smoothed reconstructions.
Overall, the advantages of \OurModel manifest in two key dimensions.
Regarding efficiency, our method achieves high-quality one-step restoration that reduces the number of function evaluations by $15\times$ to $200\times$ compared with representative multi-step methods.
Regarding effectiveness, it consistently outperforms existing one-step methods across most metrics, demonstrating that rectifying the SR trajectory via our allomorphic design effectively preserves the generative prior.

\subsubsection{Qualitative Comparisons.}

Fig.~\ref{fig:visualization} presents visual comparisons on several challenging cases.
Our method consistently produces more faithful and realistic restorations than existing one-step approaches, even surpassing some multi-step approaches.
Specifically, it effectively recovers fine-grained textures and structural details, including the intricate needlework and thread patterns on fabrics as well as the sharp contours of buildings, avoiding over-smoothed results observed in competing methods.
For text restoration, although pre-trained generative models inherently possess strong text-synthesis capability, existing one-step Real-SR methods often suffer from prior collapse during task-specific fine-tuning, resulting in truncated or distorted characters.
In contrast, \OurModel faithfully reconstructs the complete ``MARTINI'' text with clear boundaries, demonstrating superior preservation of the pre-trained generative prior.
On face restoration examples, our method accurately restores natural facial expressions and realistic dental structures.
However, many baseline methods introduce hallucinations or artifacts due to the misalignment between LR latents and generative prior.
Moreover, \OurModel successfully synthesizes high-frequency fur textures with sharp yet natural appearances, indicating the efficacy of our method in capturing complex natural image statistics.

\subsection{Ablation Study}
 
To validate the effectiveness of the core components in \OurModel, we conduct ablation studies on RealSR, DRealSR, and RealLQ250.
Specifically, we investigate the impact of the selected initial timestep, the proposed training losses, and the resolution scaling.
The quantitative results are summarized in Tab.~\ref{tab:ablation}.

\subsubsection{Justification of the Initial Timestep.}

Most existing one-step methods employ a heuristic initial timestep during both training and inference, \eg, $t=999$ in OSEDiff~\cite{wu2024one} and $t=1$ in PiSA-SR~\cite{sun2025pixel}.
However, these choices are suboptimal for Real-SR.
Specifically, within pre-trained generative prior, $t=999$ corresponds to Gaussian noise, whereas $t=1$ corresponds to a latent that is already close to the clean image.
Neither matches the latent distribution of LR images, resulting in an unnecessary distribution shift at initialization.
To validate the SNR-Guided Trajectory Initialization, we compare the selected timestep $t^*$ against these two heuristic choices.
The selected $t^*$ consistently achieves the best overall performance across both FR and NR metrics, confirming that a statistically aligned state provides an optimal initialization for one-step Real-SR.
Furthermore, by reducing the distribution shift, $t^*$ accelerates training convergence, requiring only 7k iterations compared with 9k iterations for other settings.

\subsubsection{Effectiveness of the Training Losses.}

The proposed training objectives play complementary roles in optimizing the one-step SR trajectory.
$\mathcal{L}_{\mathrm{FATC}}$ linearizes the SR flow, enabling a one-step Euler integration to better approximate the underlying probability flow, while $\mathcal{L}_{\mathrm{ATM}}$ aligns the SR flow with the allomorphic generative flow through self-adversarial distillation.
Ablating $\mathcal{L}_{\mathrm{FATC}}$ introduces severe discretization errors during one-step inference, leading to degraded perceptual quality.
Removing $\mathcal{L}_{\mathrm{ATM}}$ reveals a classic perception-distortion trade-off: while pixel-wise fidelity metrics like PSNR and SSIM improve, the perceptual metrics deteriorate substantially.
This phenomenon confirms that without the guidance of the allomorphic generative flow, the model tends to overfit the specific degradations of the training pairs, yielding over-smoothed results devoid of rich generative details.
The combination of $\mathcal{L}_{\mathrm{FATC}}$ and $\mathcal{L}_{\mathrm{ATM}}$ allows \OurModel to achieve a superior perception-distortion trade-off.

\subsubsection{Scaling to Higher Resolutions.}

Following previous one-step Real-SR methods, we train and evaluate \OurModel at a resolution of $512 \times 512$ to ensure a fair comparison in Tab.~\ref{tab:comparison}.
For the $1024 \times 1024$ images in RealLQ250, we employ a tiled inference strategy.
Since the pre-trained FLUX model is originally trained at a resolution of $1024 \times 1024$, we conduct an additional experiment by training and evaluating \OurModel at its native resolution.
The $1024 \times 1024$ setting consistently improves all metrics over the $512 \times 512$ setting.
This validates that \OurModel effectively benefits from higher-resolution training and scales well to the native resolution of the pre-trained generative prior.

\input{table/Ablation}

%% file: table/Comparison.tex
\begin{table}[t!]

\caption{Quantitative comparison among the state-of-the-art DM- and FM-based Real-SR methods on both synthetic and real-world datasets. The best and second-best results are highlighted in \colorbox{pinkhl}{\textbf{bold}} and \colorbox{purplehl}{\underline{underline}}, respectively.}
\label{tab:comparison}

\centering

\resizebox{\textwidth}{!}{
\begin{tabular}{c|ccc|ccccc}

\toprule

Methods
& StableSR \cite{wang2024exploiting}
& SeeSR \cite{wu2024seesr}
& ResShift \cite{yue2023resshift} 
& SinSR \cite{wang2024sinsr}
& OSEDiff \cite{wu2024one}
& TSD-SR \cite{dong2025tsd}
& CTMSR \cite{you2025consistency}
& \OurModel
\\

\midrule

Steps
& 200 & 50 & 15
& 1 & 1 & 1 & 1 & 1
\\

\midrule

\multicolumn{9}{c}{\textit{DIV2K-Val}}
\\

\midrule

PSNR$\uparrow$
& 23.31 & 23.71 & \cellcolor{purplehl}\underline{24.69}
& 24.43 & 23.72 & 23.02 & \cellcolor{pinkhl}\textbf{24.87} & 22.98
\\

SSIM$\uparrow$
& 0.5728 & 0.6045 & \cellcolor{purplehl}\underline{0.6175}
& 0.6012 & 0.6108 & 0.5808 & \cellcolor{pinkhl}\textbf{0.6349} & 0.6123
\\

LPIPS$\downarrow$
& 0.3129 & 0.3207 & 0.3374
& 0.3262 & 0.2941 & \cellcolor{pinkhl}\textbf{0.2673} & 0.3011 & \cellcolor{purplehl}\underline{0.2854}
\\

DISTS$\downarrow$
& 0.2138 & 0.1967 & 0.2215
& 0.2066 & 0.1976 & \cellcolor{pinkhl}\textbf{0.1821} & 0.2102 & \cellcolor{purplehl}\underline{0.1964}
\\

FID$\downarrow$
& \cellcolor{pinkhl}\textbf{24.67} & 25.83 & 36.01
& 35.45 & 26.32 & 29.16 & 25.07 & \cellcolor{purplehl}\underline{24.94}
\\

NIQE$\downarrow$
& 4.76 & 4.82 & 6.82
& 6.02 & 4.71 & \cellcolor{purplehl}\underline{4.32} & 5.3036 & \cellcolor{pinkhl}\textbf{4.18}
\\

MUSIQ$\uparrow$
& 65.63 & 68.49 & 60.92
& 62.80 & 67.97 & \cellcolor{purplehl}\underline{71.69} & 66.59 & \cellcolor{pinkhl}\textbf{72.00}
\\

MANIQA$\uparrow$
& 0.6188 & \cellcolor{purplehl}\underline{0.6239} & 0.5450
& 0.5395 & 0.6148 & 0.6192 & 0.5146 & \cellcolor{pinkhl}\textbf{0.6432}
\\

CLIPIQA$\uparrow$
& 0.6682 & \cellcolor{purplehl}\underline{0.6857} & 0.6089
& 0.6499 & 0.6683 & \cellcolor{pinkhl}\textbf{0.7416} & 0.6602 & 0.6837
\\

\midrule

\multicolumn{9}{c}{\textit{RealSR}} \\

\midrule

PSNR$\uparrow$
& 24.69 & 25.33 & \cellcolor{pinkhl}\textbf{26.31}
& \cellcolor{purplehl}\underline{26.30} & 25.15 & 24.81 & 26.00 & 24.97
\\

SSIM$\uparrow$
& 0.7052 & 0.7273 & 0.7411
& 0.7354 & 0.7341 & 0.7172 & \cellcolor{pinkhl}\textbf{0.7549} & \cellcolor{purplehl}\underline{0.7432}
\\

LPIPS$\downarrow$
& 0.3091 & 0.2985 & 0.3489
& 0.3212 & 0.2921 & \cellcolor{pinkhl}\textbf{0.2743} & 0.2905 & \cellcolor{purplehl}\underline{0.2755}
\\

DISTS$\downarrow$
& 0.2167 & 0.2213 & 0.2498
& 0.2346 & \cellcolor{purplehl}\underline{0.2128} & \cellcolor{pinkhl}\textbf{0.2104} & 0.2204 & 0.2162
\\

FID$\downarrow$
& 127.20 & 125.66 & 142.81
& 137.05 & 123.49 & \cellcolor{purplehl}\underline{114.45} & 134.97 & \cellcolor{pinkhl}\textbf{112.88}
\\

NIQE$\downarrow$
& 5.76 & 5.38 & 7.27
& 6.31 & 5.65 & \cellcolor{purplehl}\underline{5.13} & 5.60 & \cellcolor{pinkhl}\textbf{5.08}
\\

MUSIQ$\uparrow$
& 65.42 & 69.37 & 58.10
& 60.41 & 69.09 & \cellcolor{pinkhl}\textbf{71.19} & 64.52 & \cellcolor{purplehl}\underline{70.11}
\\

MANIQA$\uparrow$
& 0.6211 & \cellcolor{purplehl}\underline{0.6439} & 0.5305
& 0.5389 & 0.6326 & 0.6347 & 0.5269 & \cellcolor{pinkhl}\textbf{0.6725}
\\

CLIPIQA$\uparrow$
& 0.6195 & 0.6594 & 0.5450
& 0.6204 & 0.6693 & \cellcolor{pinkhl}\textbf{0.7160} & 0.6391 & \cellcolor{purplehl}\underline{0.6750}
\\

\midrule

\multicolumn{9}{c}{\textit{DRealSR}} \\

\midrule

PSNR$\uparrow$
& 28.04 & 28.26 & \cellcolor{purplehl}\underline{28.45}
& 28.41 & 27.92 & 27.77 & \cellcolor{pinkhl}\textbf{28.68} & 27.95
\\

SSIM$\uparrow$
& 0.7460 & 0.7698 & 0.7632
& 0.7495 & \cellcolor{pinkhl}\textbf{0.7835} & 0.7559 & \cellcolor{purplehl}\underline{0.7830} & 0.7694
\\

LPIPS$\downarrow$
& 0.3354 & 0.3197 & 0.4073
& 0.3741 & \cellcolor{purplehl}\underline{0.2968} & \cellcolor{pinkhl}\textbf{0.2967} & 0.3242 & 0.2989
\\

DISTS$\downarrow$
& 0.2287 & 0.2306 & 0.2700
& 0.2488 & 0.2165 & \cellcolor{purplehl}\underline{0.2136} & 0.2360 & \cellcolor{pinkhl}\textbf{0.2128}
\\

FID$\downarrow$
& 147.03 & 149.86 & 175.92
& 177.05 & 135.30 & \cellcolor{purplehl}\underline{134.98} & 161.24 & \cellcolor{pinkhl}\textbf{132.48}
\\

NIQE$\downarrow$
& 6.51 & 6.52 & 8.28
& 7.02 & 6.49 & \cellcolor{purplehl}\underline{5.91} & 6.21 & \cellcolor{pinkhl}\textbf{5.82}
\\

MUSIQ$\uparrow$
& 58.50 & 64.84 & 49.86
& 55.34 & 64.65 & \cellcolor{purplehl}\underline{66.62} & 59.93 & \cellcolor{pinkhl}\textbf{66.95}
\\

MANIQA$\uparrow$
& 0.5602 & 0.6026 & 0.4573
& 0.4898 & \cellcolor{purplehl}\underline{0.5899} & 0.5874 & 0.4866 & \cellcolor{pinkhl}\textbf{0.6264}
\\

CLIPIQA$\uparrow$
& 0.6171 & 0.6672 & 0.5259
& 0.6367 & \cellcolor{purplehl}\underline{0.6963} & \cellcolor{pinkhl}\textbf{0.7344} & 0.6451 & 0.6883
\\

\midrule

\multicolumn{9}{c}{\textit{RealLQ250}} \\

\midrule

NIQE$\downarrow$
& 4.62 & 4.44 & 5.33
& 5.82 & 3.97 & \cellcolor{pinkhl}\textbf{3.49} & 4.58 & \cellcolor{purplehl}\underline{3.59}
\\

MUSIQ$\uparrow$
& 63.52 & 59.27 & 57.53
& 63.73 & 69.55 & \cellcolor{purplehl}\underline{72.09} & 68.00 & \cellcolor{pinkhl}\textbf{72.51}
\\

MANIQA$\uparrow$
& \cellcolor{purplehl}\underline{0.6733} & \cellcolor{pinkhl}\textbf{0.7037} & 0.6572
& 0.5161 & 0.5782 & 0.5829 & 0.5078 & 0.6303
\\

CLIPIQA$\uparrow$
& 0.5160 & 0.7013 & 0.6129
& 0.6990 & 0.6725 & \cellcolor{pinkhl}\textbf{0.7221} & 0.6706 & \cellcolor{purplehl}\underline{0.7055}
\\

\bottomrule   
\end{tabular}
}
\end{table}

%% file: table/Ablation.tex
\begin{table*}[t!]

\caption{Ablation studies of initial timestep, training losses, and resolution scaling.}
\label{tab:ablation}
\centering
\resizebox{\textwidth}{!}{
\begin{tabular}{c|c|ccccccccc}
\toprule

Datasets
& Settings
& PSNR$\uparrow$  & SSIM$\uparrow$   & LPIPS$\downarrow$  & DISTS$\downarrow$  & FID$\downarrow$
& NIQE$\downarrow$ & MUSIQ$\uparrow$ & MANIQA$\uparrow$ & CLIPIQA$\uparrow$
\\

\midrule

\multirow{3}{*}{RealSR}
& Selected $t^*$
& 24.97 & 0.7432 & 0.2755 & 0.2162 & 112.88
& 5.08 & 70.11 & 0.6725 & 0.6750
\\

& $t = 999$
& 24.16 & 0.7385 & 0.2869 & 0.2201 & 113.62
& 5.14 & 69.75 & 0.6713 & 0.6708
\\

& $t = 1$
& 24.74 & 0.7434 & 0.2741 & 0.2156 & 112.89
& 5.27 & 68.90 & 0.6618 & 0.6681
\\

\midrule

\multirow{3}{*}{DRealSR}
& Full
& 27.95 & 0.7694 & 0.2989 & 0.2128 & 132.48
& 5.82 & 66.95 & 0.6264 & 0.6883
\\

& w/o $\mathcal{L}_{\mathrm{FATC}}$ 
& 27.88 & 0.7641 & 0.3106 & 0.2105 & 121.75
& 6.04 & 63.49 & 0.6192 & 0.6357
\\

& w/o $\mathcal{L}_{\mathrm{ATM}}$
& 28.68 & 0.7717 & 0.3008 & 0.2003 & 139.74
& 6.19 & 62.77 & 0.6049 & 0.6147
\\

\midrule

\multirow{2}{*}{RealLQ250}
& 512 $\times$ 512
& --    & --    & --    & --    & --
& 3.59  & 72.51 & 0.6303 & 0.7055
\\

& 1k $\times$ 1k
& --    & --    & --    & --    & --
& 3.52  & 72.93 & 0.6497 & 0.7105
\\

\bottomrule
\end{tabular}
}
\end{table*}

%% file: section/06-Conclusion.tex
\section{Conclusion}

In this paper, we present \OurModel, a novel FM-based framework that redefines one-step Real-SR as a specialized navigation within a generative manifold.
Inspired by the allomorphic nature between generation and restoration, our approach jointly models them within a unified flow framework.
Specifically, SNR-Guided Trajectory Initialization reduces the distribution shift between LR latents and the Gaussian prior, while Flow-Anchored Trajectory Consistency (FATC) and Allomorphic Trajectory Matching (ATM) jointly mitigate trajectory deviation and preserve the generative prior under extreme acceleration. 
Extensive experiments demonstrate that \OurModel achieves a superior balance between fidelity, realism, and efficiency, offering a promising direction toward efficient, high-quality Real-SR.

%% file: section/07-References.bib
@String(TPAMI = {IEEE Transactions on Pattern Analysis and Machine Intelligence (TPAMI)})

@String(IJCV = {International Journal of Computer Vision (IJCV)})

@String(NeurIPS = {International Conference on Neural Information Processing Systems (NeurIPS)})

@String(ICML = {International Conference on Machine Learning (ICML)})

@String(ICLR = {International Conference on Learning Representations (ICLR)})

@String(CVPR = {{IEEE/CVF} Conference on Computer Vision and Pattern Recognition (CVPR)})

@String(ICCV = {IEEE International Conference on Computer Vision (ICCV)})

@String(ECCV = {European Conference on Computer Vision (ECCV)})

@String(AAAI = {AAAI Conference on Artificial Intelligence (AAAI)})

@String(TPAMI = {TPAMI})

@String(TIP = {TIP})

@String(IJCV = {IJCV})

@String(NeurIPS = {NeurIPS})

@String(ICML = {ICML})

@String(ICLR = {ICLR})

@String(CVPR = {CVPR})

@String(ICCV = {ICCV})

@String(ECCV = {ECCV})

@String(CVPRW = {CVPRW})

@String(AAAI  = {AAAI})

@inproceedings{zhang2025uncertainty,
  author = {Leheng Zhang and Weiyi You and Kexuan Shi and Shuhang Gu},
  title = {Uncertainty-guided Perturbation for Image Super-Resolution Diffusion Model},
  booktitle = CVPR,
  year = {2025}
}

@inproceedings{dong2025tsd,
  author = {Linwei Dong and Qingnan Fan and Yihong Guo and Zhonghao Wang and Qi Zhang and Jinwei Chen and Yawei Luo and Changqing Zou},
  title = {{TSD-SR:} One-Step Diffusion with Target Score Distillation for Real-World Image Super-Resolution},
  booktitle = CVPR,
  year = {2025}
}

@inproceedings{sun2025pixel,
  author = {Lingchen Sun and Rongyuan Wu and Zhiyuan Ma and Shuaizheng Liu and Qiaosi Yi and Lei Zhang},
  title = {Pixel-level and Semantic-level Adjustable Super-resolution: {A} Dual-LoRA Approach},
  booktitle = CVPR,
  year = {2025}
}

@inproceedings{yue2025arbitrary,
  author = {Zongsheng Yue and Kang Liao and Chen Change Loy},
  title = {Arbitrary-steps Image Super-resolution via Diffusion Inversion},
  booktitle = CVPR,
  year = {2025}
}

@inproceedings{wang2024sinsr,
  author = {Yufei Wang and Wenhan Yang and Xinyuan Chen and Yaohui Wang and Lanqing Guo and Lap{-}Pui Chau and Ziwei Liu and Yu Qiao and Alex C. Kot and Bihan Wen},
  title = {SinSR: Diffusion-Based Image Super-Resolution in a Single Step},
  booktitle = CVPR,
  year = {2024}
}

@inproceedings{yin2024one,
  author = {Tianwei Yin and Micha{\"{e}}l Gharbi and Richard Zhang and Eli Shechtman and Fr{\'{e}}do Durand and William T. Freeman and Taesung Park},
  title = {One-Step Diffusion with Distribution Matching Distillation},
  booktitle = CVPR,
  year = {2024}
}

@inproceedings{wu2024seesr,
  author = {Rongyuan Wu and Tao Yang and Lingchen Sun and Zhengqiang Zhang and Shuai Li and Lei Zhang},
  title = {SeeSR: Towards Semantics-Aware Real-World Image Super-Resolution},
  booktitle = CVPR,
  year = {2024}
}

@inproceedings{meng2023distillation,
  author = {Chenlin Meng and Robin Rombach and Ruiqi Gao and Diederik P. Kingma and Stefano Ermon and Jonathan Ho and Tim Salimans},
  title = {On Distillation of Guided Diffusion Models},
  booktitle = CVPR,
  year = {2023}
}

@inproceedings{li2023lsdir,
  author = {Yawei Li and Kai Zhang and Jingyun Liang and Jiezhang Cao and Ce Liu and Rui Gong and Yulun Zhang and Hao Tang and Yun Liu and Denis Demandolx and Rakesh Ranjan and Radu Timofte and Luc Van Gool},
  title = {{LSDIR:} {A} Large Scale Dataset for Image Restoration},
  booktitle = CVPRW,
  year = {2023}
}

@inproceedings{rombach2022high,
  author = {Robin Rombach and Andreas Blattmann and Dominik Lorenz and Patrick Esser and Bj{\"{o}}rn Ommer},
  title = {High-Resolution Image Synthesis with Latent Diffusion Models},
  booktitle = CVPR,
  year = {2022}
}

@inproceedings{liang2022details,
  author = {Jie Liang and Hui Zeng and Lei Zhang},
  title = {Details or Artifacts: {A} Locally Discriminative Learning Approach to Realistic Image Super-Resolution},
  booktitle = CVPR,
  year = {2022}
}

@inproceedings{yang2022maniqa,
  author = {Sidi Yang and Tianhe Wu and Shuwei Shi and Shanshan Lao and Yuan Gong and Mingdeng Cao and Jiahao Wang and Yujiu Yang},
  title = {{MANIQA:} Multi-dimension Attention Network for No-Reference Image Quality Assessment},
  booktitle = CVPRW,
  year = {2022}
}

@inproceedings{wang2019deep,
  author = {Xintao Wang and Ke Yu and Chao Dong and Xiaoou Tang and Chen Change Loy},
  title = {Deep Network Interpolation for Continuous Imagery Effect Transition},
  booktitle  = CVPR,
  year = {2019}
}

@inproceedings{karras2019style,
  author = {Tero Karras and Samuli Laine and Timo Aila},
  title = {A Style-Based Generator Architecture for Generative Adversarial Networks},
  booktitle = CVPR,
  year = {2019}
}

@inproceedings{zhang2018unreasonable,
  author = {Richard Zhang and Phillip Isola and Alexei A. Efros and Eli Shechtman and Oliver Wang},
  title = {The Unreasonable Effectiveness of Deep Features as a Perceptual Metric},
  booktitle = CVPR,
  year = {2018}
}

@inproceedings{blau2018perception,
  author = {Yochai Blau and Tomer Michaeli},
  title = {The Perception-Distortion Tradeoff},
  booktitle = CVPR,
  year = {2018}
}

@inproceedings{ledig2017photo,
  author = {Christian Ledig and Lucas Theis and Ferenc Huszar and Jose Caballero and Andrew Cunningham and Alejandro Acosta and Andrew P. Aitken and Alykhan Tejani and Johannes Totz and Zehan Wang and Wenzhe Shi},
  title = {Photo-Realistic Single Image Super-Resolution Using a Generative Adversarial Network},
  booktitle = CVPR,
  year = {2017}
}

@inproceedings{agustsson2017ntire,
  author = {Eirikur Agustsson and Radu Timofte},
  title = {{NTIRE} 2017 Challenge on Single Image Super-Resolution: Dataset and Study},
  booktitle = CVPRW,
  year = {2017}
}

@inproceedings{you2025consistency,
  author = {Weiyi You and Mingyang Zhang and Leheng Zhang and Xingyu Zhou and Kexuan Shi and Shuhang Gu},
  title = {Consistency Trajectory Matching for One-Step Generative Super-Resolution},
  booktitle = ICCV,
  year = {2025}
}

@inproceedings{xu2025fast,
  author = {Jiaqi Xu and Wenbo Li and Haoze Sun and Fan Li and Zhixin Wang and Long Peng and Jingjing Ren and Haoran Yang and Xiaowei Hu and Renjing Pei and Pheng{-}Ann Heng},
  title = {Fast Image Super-Resolution via Consistency Rectified Flow},
  booktitle = ICCV,
  year = {2025}
}

@inproceedings{peebles2023scalable,
  author = {William Peebles and Saining Xie},
  title = {Scalable Diffusion Models with Transformers},
  booktitle = ICCV,
  year = {2023}
}

@inproceedings{zhang2021designing,
  author = {Kai Zhang and Jingyun Liang and Luc Van Gool and Radu Timofte},
  title = {Designing a Practical Degradation Model for Deep Blind Image Super-Resolution},
  booktitle = ICCV,
  year = {2021}
}

@inproceedings{ke2021musiq,
  author  = {Junjie Ke and Qifei Wang and Yilin Wang and Peyman Milanfar and Feng Yang},
  title = {{MUSIQ:} Multi-scale Image Quality Transformer},
  booktitle = ICCV,
  year = {2021}
}

@inproceedings{wang2021real,
  author = {Xintao Wang and Liangbin Xie and Chao Dong and Ying Shan},
  title = {Real-ESRGAN: Training Real-World Blind Super-Resolution with Pure Synthetic Data},
  booktitle= {ICCVW},
  year = {2021}
}

@inproceedings{liang2021swinir,
  author = {Jingyun Liang and Jiezhang Cao and Guolei Sun and Kai Zhang and Luc Van Gool and Radu Timofte},
  title = {SwinIR: Image Restoration Using Swin Transformer},
  booktitle = {ICCVW},
  year = {2021}
}

@inproceedings{cai2019toward,
  author = {Jianrui Cai and Hui Zeng and Hongwei Yong and Zisheng Cao and Lei Zhang},
  title = {Toward Real-World Single Image Super-Resolution: {A} New Benchmark and a New Model},
  booktitle = ICCV,
  year = {2019}
}

@inproceedings{zhu2017unpaired,
  author = {Jun{-}Yan Zhu and Taesung Park and Phillip Isola and Alexei A. Efros},
  title = {Unpaired Image-to-Image Translation Using Cycle-Consistent Adversarial Networks},
  booktitle = ICCV,
  year = {2017}
}

@inproceedings{lin2024diffbir,
  author= {Xinqi Lin and Jingwen He and Ziyan Chen and Zhaoyang Lyu and Bo Dai and Fanghua Yu and Yu Qiao and Wanli Ouyang and Chao Dong},
  title = {DiffBIR: Toward Blind Image Restoration with Generative Diffusion Prior},
  booktitle = ECCV,
  year = {2024}
}

@inproceedings{yang2024pixel,
  author = {Tao Yang and Rongyuan Wu and Peiran Ren and Xuansong Xie and Lei Zhang}, 
  title = {Pixel-Aware Stable Diffusion for Realistic Image Super-Resolution and Personalized Stylization},
  booktitle = ECCV,
  year = {2024}
}

@inproceedings{zhang2022perception,
  author = {Yuehan Zhang and Bo Ji and Jia Hao and Angela Yao},
  title = {Perception-Distortion Balanced {ADMM} Optimization for Single-Image Super-Resolution},
  booktitle = ECCV,
  year = {2022}
}

@inproceedings{wei2020component,
  author = {Pengxu Wei and Ziwei Xie and Hannan Lu and Zongyuan Zhan and Qixiang Ye and Wangmeng Zuo and Liang Lin},
  title = {Component Divide-and-Conquer for Real-World Image Super-Resolution},
  booktitle = ECCV,
  year = {2020}
}

@inproceedings{dong2014learning,
  author = {Chao Dong and Chen Change Loy and Kaiming He and Xiaoou Tang},
  title = {Learning a Deep Convolutional Network for Image Super-Resolution},
  booktitle = ECCV,
  year = {2014}
}

@inproceedings{wu2024one,
  author = {Rongyuan Wu and Lingchen Sun and Zhiyuan Ma and Lei Zhang},
  title = {One-Step Effective Diffusion Network for Real-World Image Super-Resolution},
  booktitle = NeurIPS,
  year = {2024}
}

@inproceedings{yin2024improved,
  author = {Tianwei Yin and Micha{\"{e}}l Gharbi and Taesung Park and Richard Zhang and Eli Shechtman and Fr{\'{e}}do Durand and William Freeman},
  title = {Improved Distribution Matching Distillation for Fast Image Synthesis},
  booktitle = NeurIPS,
  year = {2024}
}

@inproceedings{ai2024dreamclear,
  author = {Yuang Ai and Xiaoqiang Zhou and Huaibo Huang and Xiaotian Han and Zhengyu Chen and Quanzeng You and Hongxia Yang},
  title = {DreamClear: High-Capacity Real-World Image Restoration with Privacy-Safe Dataset Curation},
  booktitle = NeurIPS,
  year = {2024}
}

@inproceedings{yue2023resshift,
  author = {Zongsheng Yue and Jianyi Wang and Chen Change Loy},
  title = {ResShift: Efficient Diffusion Model for Image Super-resolution by Residual Shifting},
  booktitle = NeurIPS,
  year = {2023}
}

@inproceedings{wang2023prolificdreamer,
  author = {Zhengyi Wang and Cheng Lu and Yikai Wang and Fan Bao and Chongxuan Li and Hang Su and Jun Zhu}, 
  title = {ProlificDreamer: High-Fidelity and Diverse Text-to-3D Generation with Variational Score Distillation},
  booktitle = NeurIPS,
  year = {2023}
}

@inproceedings{ho2020denoising,
  author = {Jonathan Ho and Ajay Jain and Pieter Abbeel},
  title = {Denoising Diffusion Probabilistic Models},
  booktitle = NeurIPS,
  year = {2020}
}

@inproceedings{song2019generative,
  author = {Yang Song and Stefano Ermon},
  title = {Generative Modeling by Estimating Gradients of the Data Distribution},
  booktitle = NeurIPS,
  year = {2019}
}

@inproceedings{heusel2017gans,
  title={GANs trained by a two time-scale update rule converge to a local nash equilibrium},
  author={Heusel, Martin and Ramsauer, Hubert and Unterthiner, Thomas and Nessler, Bernhard and Hochreiter, Sepp},
  booktitle=NeurIPS,
  year={2017}
}

@inproceedings{goodfellow2014generative,
  author = {Ian J. Goodfellow and Jean Pouget{-}Abadie and Mehdi Mirza and Bing Xu and David Warde{-}Farley and Sherjil Ozair and Aaron C. Courville and Yoshua Bengio}, 
  title = {Generative Adversarial Nets},
  booktitle = NeurIPS,
  year = {2014}
}

@inproceedings{song2023consistency,
  author = {Yang Song and Prafulla Dhariwal and Mark Chen and Ilya Sutskever}, 
  title = {Consistency Models},
  booktitle = ICML,
  year = {2023}
}

@inproceedings{nichol2021improved,
  author = {Alexander Quinn Nichol and Prafulla Dhariwal}, 
  title = {Improved Denoising Diffusion Probabilistic Models},
  booktitle = ICML,
  year = {2021}
}

@inproceedings{cheng2026twinflow,
  author = {Zhenglin Cheng and Peng Sun and Jianguo Li and Tao Lin},
  title = {TwinFlow: Realizing One-step Generation on Large Models with Self-adversarial Flows},
  booktitle = ICLR,
  year = {2026}
}

@inproceedings{kim2024consistency,
  author = {Dongjun Kim and Chieh{-}Hsin Lai and Wei{-}Hsiang Liao and Naoki Murata and Yuhta Takida and Toshimitsu Uesaka and Yutong He and Yuki Mitsufuji and Stefano Ermon},
  title = {Consistency Trajectory Models: Learning Probability Flow {ODE} Trajectory of Diffusion},
  booktitle = ICLR,
  year = {2024}
}

@inproceedings{liu2024instaflow,
  author = {Xingchao Liu and Xiwen Zhang and Jianzhu Ma and Jian Peng and Qiang Liu},
  title = {InstaFlow: One Step is Enough for High-Quality Diffusion-Based Text-to-Image Generation},
  booktitle = ICLR,
  year = {2024}
}

@inproceedings{lipman2023flow,
  author = {Yaron Lipman and Ricky T. Q. Chen and Heli Ben{-}Hamu and Maximilian Nickel and Matthew Le},
  title = {Flow Matching for Generative Modeling},
  booktitle = ICLR,
  year = {2023}
}

@inproceedings{liu2022flow,
  author = {Xingchao Liu and Chengyue Gong and Qiang Liu},
  title = {Flow Straight and Fast: Learning to Generate and Transfer Data with Rectified Flow},
  booktitle = ICLR,
  year = {2023}
}

@inproceedings{albergo2023building,
  author = {Michael S. Albergo and Eric Vanden{-}Eijnden},
  title = {Building Normalizing Flows with Stochastic Interpolants},
  booktitle = ICLR,
  year = {2023}
}

@inproceedings{salimans2022progressive,
  author = {Tim Salimans and Jonathan Ho},
  title = {Progressive Distillation for Fast Sampling of Diffusion Models},
  booktitle = ICLR,
  year = {2022}
}

@inproceedings{hu2022lora,
  author = {Edward J. Hu and Yelong Shen and Phillip Wallis and Zeyuan Allen{-}Zhu and Yuanzhi Li and Shean Wang and Lu Wang and Weizhu Chen},
  title = {LoRA: Low-Rank Adaptation of Large Language Models},
  booktitle = ICLR,
  year = {2022}
}

@inproceedings{song2021score,
  author = {Yang Song and Jascha Sohl{-}Dickstein and Diederik P. Kingma and Abhishek Kumar and Stefano Ermon and Ben Poole},
  title = {Score-Based Generative Modeling through Stochastic Differential Equations},
  booktitle = ICLR,
  year = {2021}
}

@inproceedings{kingma2014adam,
  author = {Diederik P. Kingma and Jimmy Ba},
  title = {Adam: {A} Method for Stochastic Optimization},
  booktitle = ICLR,
  year = {2015}
}

@inproceedings{wang2023exploring,
  author = {Jianyi Wang and Kelvin C. K. Chan and Chen Change Loy},
  title = {Exploring {CLIP} for Assessing the Look and Feel of Images},
  booktitle = AAAI,
  year = {2023}
}

@article{dong2016image,
  author = {Chao Dong and Chen Change Loy and Kaiming He and Xiaoou Tang},
  title = {Image Super-Resolution Using Deep Convolutional Networks},
  journal = TPAMI,
  year = {2016}
}

@article{wang2024exploiting,
  author = {Jianyi Wang and Zongsheng Yue and Shangchen Zhou and  Kelvin C. K. Chan and Chen Change Loy},
  title = {Exploiting Diffusion Prior for Real-World Image Super-Resolution},
  journal = IJCV,
  year = {2024}
}

@article{ding2020image,
  author = {Keyan Ding and Kede Ma and Shiqi Wang and Eero P. Simoncelli},
  title = {Image Quality Assessment: Unifying Structure and Texture Similarity},
  journal = TPAMI,
  year = {2022}
}

@article{zhang2015feature,
  author = {Lin Zhang and Lei Zhang and Alan C. Bovik},
  title = {A Feature-Enriched Completely Blind Image Quality Evaluator},
  journal = TIP,
  year = {2015}
}

@article{wang2004image,
  author = {Zhou Wang and Alan C. Bovik and Hamid R. Sheikh and Eero P. Simoncelli},
  title = {Image quality assessment: from error visibility to structural similarity},
  journal = TIP,
  year = {2004}
}

@article{albergo2025stochastic,
  author = {Michael S. Albergo and Nicholas M. Boffi and Eric Vanden{-}Eijnden},
  title = {Stochastic Interpolants: {A} Unifying Framework for Flows and Diffusions},
  journal = {JMLR},
  year = {2025}
}

@article{wu2025omgsr,
  author = {Zhiqiang Wu and Zhaomang Sun and Tong Zhou and Bingtao Fu and Ji Cong and Yitong Dong and Huaqi Zhang and Xuan Tang and Mingsong Chen and Xian Wei},
  title  = {{OMGSR:} You Only Need One Mid-timestep Guidance for Real-World Image Super-Resolution},
  journal = {arXiv:2508.08227},
  year={2025}
}

@article{yang2024consistency,
  author = {Ling Yang and Zixiang Zhang and Zhilong Zhang and Xingchao Liu and Minkai Xu and Wentao Zhang and Chenlin Meng and Stefano Ermon and Bin Cui},
  title = {Consistency Flow Matching: Defining Straight Flows with Velocity Consistency},
  journal = {arXiv:2407.02398},
  year={2024}
}

@article{luo2023latent,
  author = {Simian Luo and Yiqin Tan and Longbo Huang and Jian Li and Hang Zhao},
  title = {Latent Consistency Models: Synthesizing High-Resolution Images with Few-Step Inference},
  journal = {arXiv:2310.04378},
  year={2023}
}
